\pdfoutput=1

\documentclass[11pt]{article}

\usepackage[]{acl}

\usepackage{times}
\usepackage{latexsym}

\usepackage{amsmath,amsfonts,bm}
\usepackage{graphicx}
\usepackage{multirow}
\usepackage{makecell}
\usepackage{array}
\usepackage[para]{threeparttable}
\usepackage{booktabs}
\usepackage{tabularx}
\usepackage{xspace}
\usepackage{caption}
\usepackage{subcaption}
\usepackage{longtable}
\usepackage{hyperref}
\usepackage{bbm}
\usepackage{bm}
\usepackage{enumitem}
\usepackage{algorithm}
\usepackage{algorithmic}
\usepackage{geometry}
\usepackage{comment}
\usepackage[colorinlistoftodos]{todonotes}
\usepackage{tablefootnote}
\usepackage{color}
\usepackage{tcolorbox}
\usepackage{tikz}
\usepackage{tabularray}
\usepackage[normalem]{ulem}

\usepackage{listings}
\usepackage{xcolor}
\usepackage{xspace}
\usepackage[utf8]{inputenc}
\usepackage[T1]{fontenc}

\usepackage{microtype}
\usepackage{inconsolata}

\definecolor{main}{HTML}{5989cf}    
\definecolor{sub}{HTML}{cde4ff}    

\tcbset{
    sharp corners,
    colback = white,
    before skip = 0.2cm,   
    after skip = 0.5cm     
}    

\newtcolorbox{boxC}{
    colback = sub, 
    boxrule = 0pt 
}

\usepackage{graphicx}

\newcommand{\Ours}{\textsc{Context organizer}\xspace}
\newcommand{\ours}{\textsc{context organizer}\xspace}
\newcommand{\framework}{\textsc{COrg}\xspace}

\newcommand{\conflictqa}{ConflictQA\xspace}
\newcommand{\ambig}{AmbigDocs\xspace}
\newcommand{\ambigplus}{AmbigDocs+\xspace}
\newcommand{\conflictqaplus}{ConflictQA+\xspace}

\title{\framework: Generating Answers from Complex, Interrelated Contexts}

\author{
  Hyunji Lee$^{\hspace{.1em}{\boldsymbol{\kappa}}}$\thanks{\hspace{1mm}Work performed during internship at Adobe Research.} \quad
  Franck Dernoncourt$^{\hspace{.1em}\boldsymbol{\alpha}}$ \quad
  Trung Bui$^{\hspace{.1em}\boldsymbol{\alpha}}$ \quad
  Seunghyun Yoon$^{\hspace{.1em}\boldsymbol{\alpha}}$ \quad \\
    \\
  $^{\kappa\hspace{.1em}}$KAIST AI \quad $^{\alpha\hspace{.1em}}$Adobe Research\quad \\
   \texttt  {hyunji.amy.lee@kaist.ac.kr} \quad \texttt {\{dernonco, bui, syoon\}@adobe.com}\\
}

\begin{document}
\maketitle 

\begin{abstract}

In a real-world corpus, knowledge frequently recurs across documents but often contains inconsistencies due to ambiguous naming, outdated information, or errors, leading to complex interrelationships between contexts. Previous research has shown that language models struggle with these complexities, typically focusing on single factors in isolation. We classify these relationships into four types: distracting, ambiguous, counterfactual, and duplicated. Our analysis reveals that no single approach effectively addresses all these interrelationships simultaneously. Therefore, we introduce \ours~(\framework), a framework that organizes multiple contexts into independently processed groups. This design allows the model to efficiently find all relevant answers while ensuring disambiguation. \framework consists of three key components: a graph constructor, a reranker, and an aggregator. Our results demonstrate that \framework balances performance and efficiency effectively, outperforming existing grouping methods and achieving comparable results to more computationally intensive, single-context approaches. 

\end{abstract}

\section{Introduction}

\begin{figure}[t!]
    \centering
    \begin{minipage}[b]{0.45\textwidth}
    \includegraphics[width=\textwidth]{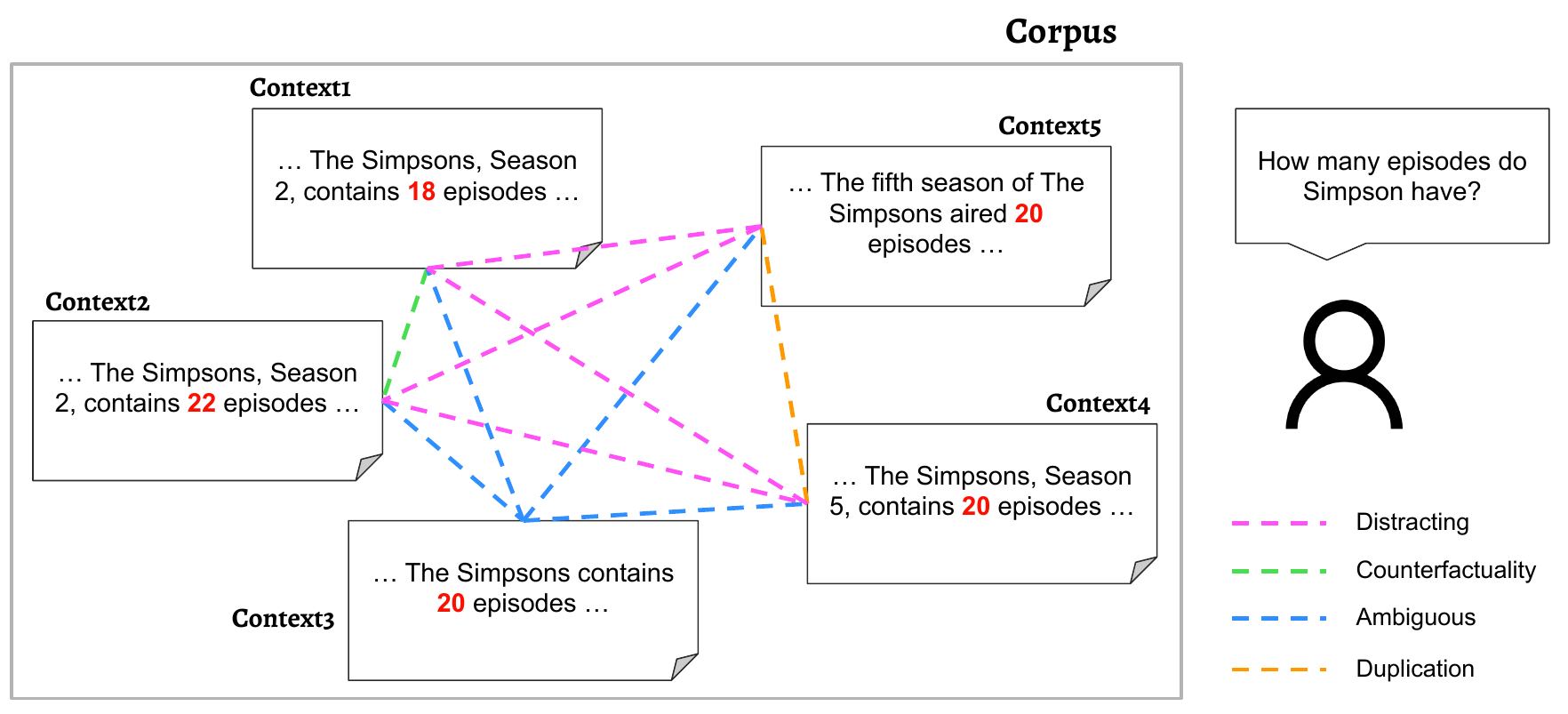}
    \end{minipage}
\caption{\fontsize{6.5}{10}\footnotesize In real-world corpora, contexts often exhibit complex interrelationships, which we classify into four categories: distracting, counterfactual, duplicated, and ambiguous.}
\label{fig: fig1}
\end{figure}

In real-world documents—ranging from blog posts and news articles to official records or user-generated content—the same knowledge often appears in multiple forms, sometimes with consistency but often with variations or conflicts. These discrepancies can arise from ambiguous phrasing, outdated data, or simple errors. When analyzing these contexts, we find relationships between them generally fall into four categories as shown in Figure~\ref{fig: fig1}: \textit{distracting, ambiguous, counterfactual, or duplicated}. Each entity consists of a surface name, a general term that can be ambiguous, and a descriptor that provides specificity. For instance, in ``The Simpsons (Season 2) contains 22 episodes'', ``The Simpsons'' is the surface name, and ``Season 2'' is the descriptor that specifies the entity. Based on these attributes, we classify contexts as distracting (same surface name, different descriptors), ambiguous (same surface name, only one with a descriptor), counterfactual (same entity with differing answers), or duplicated (same entity with identical answers).

Current research often simplifies real-world complexities by treating corpora as unified sources (e.g., fixed Wikipedia versions), where knowledge appears consistently without cross-document conflicts. Many studies also address isolated factors rather than considering such complex interrelationships between contexts~\citep{min-etal-2020-ambigqa, lee2024ambigdocs, xu2024knowledge, Zhang2021SituatedQAIE}. Additionally, finding relevant contexts using web-based retrieval methods often introduces search engine biases, resulting in limited diversity~\citep{lee2024routerretriever, gezici2021evaluation}.
To bridge this gap, we expand existing corpora to better reflect such complex interrelation between contexts in real-world conditions.
Specifically, we introduce \ambigplus and \conflictqaplus—extensions of AmbigDocs~\citep{lee2024ambigdocs} and ConflictQA~\citep{Zhou2023ContextfaithfulPF}—where we construct additional contexts based on the (question, answer, context) pairs, ensuring coverage of all four conditions.

When analyzing the effect of each factor, we observe a performance drop, particularly when ambiguous or counterfactual contexts are added, consistent with prior findings~\citep{lee2024ambigdocs, Zhou2023ContextfaithfulPF, lee2023well}. Our investigation reveals that no single simple solution addresses such contexts in complex relationships simultaneously; although simple methods exist for individual factors, they often fail to generalize across scenarios.
Interestingly, we found that a straightforward solution for distracting contexts is to modify the question to a plural form. For ambiguous contexts, adding or replacing missing descriptors to create a distracting relationship can improve clarity. However, these approaches are less effective for counterfactual contexts, where separating them into different forward passes appears to yield better performance.

To address these challenges, we introduce \Ours~(\framework), a framework designed to improve performance on real-world corpora through a simple, efficient approach based on insights from individual solutions. \framework prioritizes three objectives: (1) high answer recall, (2) accurate disambiguation, and (3) minimal inference runs. For cases with multiple answers, \framework generates responses that include all relevant answers with citations, allowing users to review and filter information as needed. \framework comprises three components: the \textbf{graph constructor}, which represents context relationships; the \textbf{reranker}, which organizes contexts into scenario-based groups; and the \textbf{aggregator}, which generates responses with citations for each group.

We conduct experiments using eight language models of different sizes and observe that \ours consistently improves performance over six baselines on both \ambigplus and \conflictqaplus, which contain multiple factors, as well as on \ambig and \conflictqa, which each contain only a single factor. \ours notably excels in entity recall, measuring the model’s ability to identify disambiguated entities. Even large models show low disambiguation performance when simply processed without \framework. Additionally, grouping similar contexts without structured processing, as in \framework, tends to reduce performance: groups with only similar contexts seem to confuse the model more than simply appending diverse contexts together. We hope our analysis of these real-world corpus scenarios encourages the community to explore the unique effects and solutions for each factor in greater depth.

\section{Complex, Interrelated Contexts} \label{sec: factor}

We analyze real-world contexts and observe that the relationship between contexts can be categorized into four types: \textit{distracting, ambiguous, counterfactual, and duplicated}. In Section~\ref{sec3: def}, we define these four categories, followed by an analysis of their occurrence in real-world web corpora in Section~\ref{sec3: bing}. In Section~\ref{sec3: construction}, we describe how we extend an existing dataset to incorporate all four relationship types, and in Section~\ref{sec3: metric}, we share details of evaluation metrics.

\subsection{Definition of Relationships} \label{sec3: def}

When given a corpus \( \mathcal{C} = \{c_1, \cdots, c_N\} \), where each context $c_i$ consists of multiple sentences, we define the \text{relationship between the contexts} by breaking down the information in each sentence into three components: surface name \( s_i \), descriptor \( d_i \), and answer \( a_i \). 
For example, in Doc2 of Figure~\ref{fig: fig1}, the sentence “The Simpsons, Season 2, contains 22 episodes” is parsed as follows: “The Simpsons” is the surface name (a general, potentially ambiguous entity), “Season 2” is the descriptor (specific detail that disambiguates between entities with the same surface name), and “22 episodes” is the answer. 

An answer exists only when the context is \textit{relevant} to the question; otherwise, the answer is considered null. 
To determine relevance between question and context, given a question about an entity \( e_q \) and descriptor \( d_q \) with the corpus \( \mathcal{C} \), if \( d_q \) is null, relevant contexts include all contexts where \( e_i = e_q \). If \( d_q \) is not null, relevant contexts are limited to those where both \( e_i = e_q \) and \( d_i = d_q \).

When given a question and two contexts relevant to the question, \( c_i \) and \( c_j \), we can extract information \((e_i, d_i, a_i)\) and \((e_j, d_j, a_j)\) from each context where $e_q = e_i = e_j$. Based on the extracted info, the four relationships between contexts are defined as:

\begin{itemize}
    \item \textbf{Ambiguous case}: \( d_i \neq d_j \) with either $d_i = Null$ or $d_j = Null$
    \item \textbf{Distracting case}: \( d_i \neq d_j \) with $d_i \neq Null$ and $d_j \neq Null$
    \item \textbf{Counterfactual case}: \( d_i = d_j \) and \( a_i \neq a_j \).
    \item \textbf{Duplicated case}: \( d_i = d_j \) and \( a_i = a_j \).
\end{itemize}
\noindent where $Null$ indicates that it is empty.

\subsection{Statistics of Real-World Corpora}  \label{sec3: bing}

To understand the structure of real-world corpora, we analyze the relationship between the top 10 contexts retrieved for questions from AmbigDocs~\citep{lee2024ambigdocs} using the Bing API\footnote{https://www.microsoft.com/en-us/bing/apis/bing-web-search-api}. Among these, we found an average composition of 25.2\% ambiguous, 34.7\% duplicated, 12.4\% conflicting, and 27.7\% distracting relationship, underscoring the need to address all four factors together rather than in isolation.
Moreover, only 32.7\% of the diverse answers from AmbigDocs questions were covered, indicating a potential retrieval bias in search engines~\citep{lee2024routerretriever, gezici2021evaluation}. To avoid this limitation and ensure a balanced representation of all factors, we extend a corpus instead of relying solely on web-crawled contexts, with further details provided in the next section. For details on how the statistics were obtained, see Appendix~\ref{app: stats}.

\begin{table}[t!]
\centering
\fontsize{8.5}{10}\selectfont
    \begin{tabular}{c|cc|cc}
    \toprule
    & \multicolumn{2}{c}{\ambig} & \multicolumn{2}{c}{\conflictqa} \\
    \midrule
    & Original & New (+) & Original & New (+) \\
    \midrule
    Ambiguous & \textbf{N} & Y & \textbf{N} & Y \\ 
    Distracting & Y & Y & \textbf{N} & Y \\
    Conflicting & \textbf{N} & Y & Y & Y \\
    Duplicated & \textbf{N} & Y & \textbf{N} & Y\\
    \bottomrule
    \end{tabular}
\caption
     {\fontsize{8.5}{10}\footnotesize Overview of datasets: \ambig and \conflictqa include a single factor, while \ambigplus and \conflictqaplus incorporate all four factors.} 
\label{table: dataset}
\end{table}

\subsection{Corpus Construction} \label{sec3: construction}

To evaluate LLMs in real-world scenarios, we expand existing datasets with additional contexts based on (question, answer, contexts) pairs so that the corpus with related contexts for each question contains all four factors. We use \ambig~\citep{lee2024ambigdocs}, which includes distracting contexts, and a dataset from \citet{Zhou2023ContextfaithfulPF}, which we call \conflictqa, containing counterfactual contexts (Table~\ref{table: dataset}). We construct extended versions, \ambigplus and \conflictqaplus, by adding ambiguous, conflicting, and duplicated contexts to \ambig and distracting, ambiguous, and duplicated contexts to \conflictqa.

Specifically, \ambig provides a question with related contexts where the question references an entity without a specific descriptor. 
Each context pair consists of a sub-question with a descriptor and corresponding answer. 
To add counterfactual contexts, we select a sub-question and for each answer in pairs, instruct GPT-4~\citep{achiam2023gpt} to generate contexts. The duplicated and ambiguous contexts are generated by providing the model with the answer with a sub-question or question, respectively.
For \conflictqa, which includes questions with counterfactual contexts often lacking descriptors, we use GPT-4 to generate plausible sub-questions with descriptors to match a similar format with \ambig. Distracting contexts are created for each sub-question and answer, while duplicated and ambiguous contexts follow the same process as in \ambig.

After generating the corpus, we apply two filtering processes: (1) inclusion of answer in the generated context and (2) whether GPT-4o answers sub-questions correctly when given the context. If either fails, we regenerate with GPT-4 with the issue added until it passes both filters.
We then hire five freelancers to evaluate a random 10\% sample of \ambigplus, assessing whether (1) generated contexts are relevant to the question, (2) answers are accurate and present within the document, and (3) the corpus represents the expected variety of real-world context relationships. We achieve average ratings of 93.4\%, 89.0\%, and 84.6\% across each criterion, indicating high quality. Further details on calculation methods, context generation, and dataset statistics are provided in Appendix~\ref{app: sec2}.

\subsection{Evaluation Metric}
\label{sec3: metric}

Following \citet{lee2024ambigdocs}, we assess the model on AmbigDocs with four metrics. 
\textit{Entity Recall (Ent)}, calculates the average token-level recall for a descriptor. 
\textit{Answer Recall (Ans)}, measures the average token-level recall for each correct answer. 
\textit{Entity-Answer Recall (EAR)} averages the product of entity recall and answer recall to measure how well the model generates both answers with their corresponding descriptor. 
\textit{Disambig-F1 (D-F1)}~\citep{stelmakh2022asqa} is a model-based metric that assesses answer recall by comparing the model’s response to the correct answer; the response is generated by a RoBERTa-based QA model trained on SQuAD-v2, with sub-questions as input. 
For ConflictQA, we report only Answer Recall (Ans) and Disambig-F1 (D-F1), as descriptors for each context are not labeled.

\section{Analyzing Solution for each Factor}
In this section, we analyze how adding the context of each factor to the input affects model performance~(Section~\ref{sec4: factor}) and investigate a simple solution when looking at each factor individually (Section~\ref{sec4: solve}). All evaluation in this section is performed with the Llama2 7B model~\citep{touvron2023llama}.

\subsection{Affect of each factor} \label{sec4: factor}

\begin{figure}[t!]
    \centering
    \begin{minipage}[b]{0.45\textwidth}
    \includegraphics[width=\textwidth]{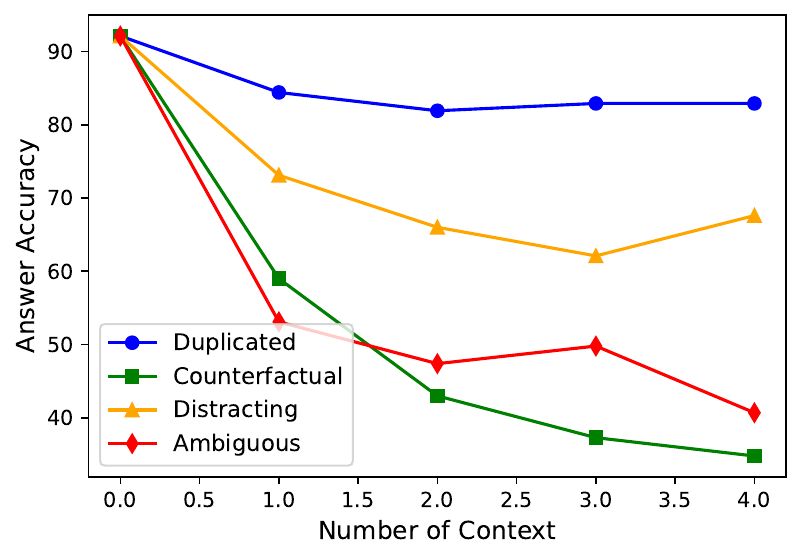}
    \end{minipage}
\caption{\fontsize{6.5}{10}\footnotesize Answer recall as the number of contexts increases for each factor. For ambiguous cases, note that since ambiguity can only exist between two contexts, sets with three or more contexts (x > 2) also include distracting relationships.}
\label{fig: overall}
\end{figure}

To assess the impact of each factor, we sample 1k instances from \ambigplus. For each question, we analyze how adding context to each relationship affects overall model performance. Figure~\ref{fig: overall} presents the performance trends as contexts reflecting each factor are added.
Our analysis reveals that adding duplicated contexts has minimal impact on overall performance. Introducing distracting contexts results in a slight performance drop, though not as significant as it is easy for LLM to distinguish entities when they have different descriptors. However, the inclusion of counterfactual and ambiguous contexts leads to the most substantial performance degradation. These contexts typically cause the model to generate answers that cover only a subset of possible responses, rather than providing a comprehensive set of answers.

\subsection{Solution for Each Factors individually}  \label{sec4: solve}
In this section, we investigate solutions for each factor individually based on the observation in the above section.

\paragraph{Distracting Context}
\begin{table}[t!]
\centering
\fontsize{8.5}{10}\selectfont
    \begin{tabular}{c|cccc}
    \toprule
    & Ent & Ans & EAR & D-F1 \\
    \midrule
    & 52.5 & 53.0 & 36.8 & 21.7 \\
    One Shot & 48.4 & 61.4 & 37.6 & 18.2 \\
    Extra Prompt & 53.0 &56.3& 38.3 & 27.6 \\
    \textbf{Plural} & 68.9 & 67.9 & 42.7 & 28.0 \\
    \bottomrule
    \end{tabular}
\caption
     {\fontsize{8.5}{10}\footnotesize Performance of Llama2-7B using only contexts with distracting relationships. Changing the question to a plural format shows the highest improvement.} 
\label{table: distracting}
\end{table}

Previous works~\cite{Zhang2021SituatedQAIE, lee2024ambigdocs} have shown that models often struggle to answer questions with multiple distracting contexts, typically selecting just one answer instead of generating all possible answers. 
To address this, we explore three approaches to inform the model about multiple answers: (1) adding a prompt\footnote{The question may be ambiguous, thereby containing multiple answers} indicating that the question may be ambiguous and could have multiple answers, (2) providing an example in the input, and (3) changing the question's tense from singular to plural\footnote{``2019 World Ice Hockey Championships host country?'' $\rightarrow$ ``2019 World Ice Hockey Championships host countries?''}. Surprisingly, pluralizing the question yielded the greatest improvement, as shown in Table~\ref{table: distracting}\footnote{Simply pluralizing every question is ineffective; when only a single context is present, plural forms often lead the model to generate multiple answers unintentionally.}. This was followed by the additional prompt and the one-shot example, though the latter sometimes degraded performance, likely due to the model drawing on knowledge from the example. 
However, despite these enhancements, the model still struggles to consistently generate all possible answers and disambiguate between them and their corresponding entities.

\paragraph{Ambiguous Context}

\begin{table}[t!]
\centering
\fontsize{8.5}{10}\selectfont
    \begin{tabular}{c|cccc}
    \toprule
    & Ent & Ans & EAR & D-F1 \\
    \midrule
    & 31.0 & 49.6 & 29.9 & 17.2 \\
    One Shot & 34.4 & 52.2 & 32.0 & 18.3 \\
    Extra Prompt & 40.7 & 50.9 & 31.5 & 18.4 \\
    Plural & 45.1 & 51.9 & 33.2 & 20.1 \\
    \textbf{Change to Distracting} & \text{52.5} & \text{53.0} & \text{36.8} & \text{21.7} \\
    \bottomrule
    \end{tabular}
\caption
     {\fontsize{8.5}{10}\footnotesize Performance of Llama2-7B using only contexts with ambiguous relationships. Changing the relationship to distracting one shows the highest performance.} 
\label{table: ambiguous}
\end{table}

While the difference between distracting and ambiguous contexts is minor, whether context with an empty descriptor is included, their impact varies significantly; distracting contexts generally show less performance degradation than ambiguous ones. 
Thus, as shown in Table~\ref{table: ambiguous}, we conducted experiments where when contexts with ambiguous relationships are given, we replace the context with empty distractor to another context that has the same content but with a descriptor\footnote{Please note due to data construction method of \ambigplus, all contexts without descriptor is replaceable to corresponding document with descriptor, but we also consider the case where it is irreplaceable, which we further discuss in Section~\ref{ours}.}. In other words, shifting ambiguous relationships to distracting ones. 
This replacement method (Change to Distracting) yields the highest performance, outperforming other approaches like providing examples, prompts, or rephrasing questions in plural format.

\paragraph{Counterfactual Contexts}

\begin{table}[t!]
\centering
\fontsize{8.5}{10}\selectfont
    \begin{tabular}{c|cccc}
    \toprule
    & Ent & Ans & EAR & D-F1\\
    \midrule
    & 30.1 & 44.6 & 28.6 & 18.0\\
    One Shot & 36.4 & 51.0 & 32.6 & 20.4\\
    Extra Prompt & 29.4 & 45.2 & 27.1 & 18.7\\
    Plural & 34.8 & 50.5 & 30.7 & 19.8\\
    \textbf{Separation} & 51.2 & 85.9 & 40.8 & 26.1\\
    \bottomrule
    \end{tabular}
\caption
     {\fontsize{8.5}{10}\footnotesize Performance of Llama2-7B with only contexts in counterfactual relationship. Separation yields the highest performance, while other cases show small improvement.} 
\label{table: counter}
\end{table}

Previous works indicate that language models tend to favor contexts aligning with their parametric knowledge when presented with multiple counterfactual contexts~\citep{Chen2022RichKS, Xie2023AdaptiveCO, lee2023well}, particularly struggling as the number of conflicting contexts grows~\citep{Jin2024TugofWarBK}. We first test whether solutions effective for distracting or ambiguous contexts also improve performance in counterfactual contexts. Results in Table~\ref{table: counter} show that these solutions yield smaller improvements in counterfactual cases. While prior studies propose heavy pipelines to resolve such conflicting cases~\citep{Wu2022TopologicalAO, Hsu2021WikiContradictionDS}, as our approach aims to address not only conflicting case but all four context types together; thus, we explore a simple method by processing each context individually, observing significant gains with separated contexts.

\begin{figure*}[t!]
    \centering
    \begin{minipage}[b]{1.0\textwidth}
    \includegraphics[width=\textwidth]{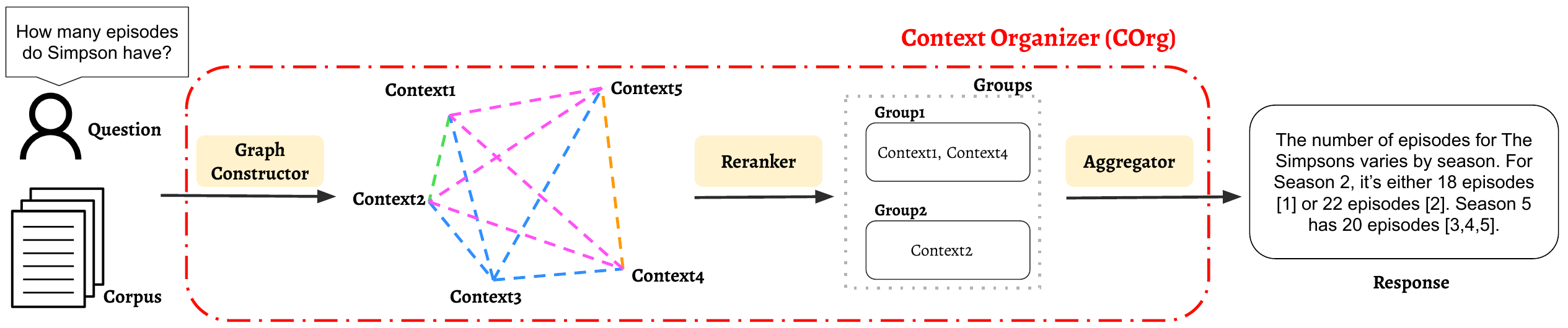}
    \end{minipage}
\caption{\fontsize{6.5}{10}\footnotesize Overview of \ours~(\framework), composed of three components: graph constructor, reranker, and aggregator. Given a corpus with multiple relevant contexts and a question, the graph constructor and the reranker organize the corpus based on the question, and the aggregator generates a response containing all possible answers with references.}
\label{fig: overall}
\end{figure*}

\paragraph{Duplicated Contexts}
The results in Figure~\ref{fig: overall} indicate that adding duplicated contexts generally does not impact overall performance. 
However, we observe a slight performance drop, likely due to a longer input. Thus, for both performance and efficiency, retaining only a single instance of duplicated contexts seems to be a good approach. The choice of which duplicated context to keep does not seem to make a notable difference.

\section{\Ours} \label{ours}

In this section, we introduce \ours (\framework), a framework designed for real-world corpora, based on observations in Section~\ref{sec4: solve}. As shown in Figure~\ref{fig: overall}, \framework consists of three components: a graph constructor, a reranker, and an aggregator. The framework aims to achieve (1) high answer recall, (2) strong disambiguation, and (3) efficiency through reduced inference runs.

\paragraph{Graph Constructor} 
The \textit{Graph Constructor} component constructs a graph from a list of contexts, where each node represents a context and each edge indicates the relationship between contexts. We employ GPT-4 to identify these relationships. To efficiently capture all relationships\footnote{Here, we distinguish \textit{ambiguous} into two types: whether a context contains the same information as another context (and is therefore interchangeable) or not, allowing us to apply the solution outlined in Section~\ref{sec4: solve}. We consider contexts that are interchangeable as having an "ambiguous" relationship. Further details are provided in Appendix~\ref{app: ours}.}, we use an iterative approach as in Algorithm~\ref{alg:graph_construction}.
Initially, we extract relationships between context 1 and the remaining contexts (line 6-10 in Algorithm~\ref{alg:graph_construction}). If two contexts are classified as counterfactual or duplicated, their relationships with other contexts will mirror each other (line 12-16). For instance, if context 1 and context 2 are counterfactual, then context 2's relationships with the remaining contexts will reflect context 1’s relationships. For other relationships, this mirroring only applies in counterfactual or duplicated cases (lines 17–25). Such nodes which have missing edges are processed in subsequent iterations. This approach minimizes redundant checks and reduces the number of iterations by focusing only on missing edges.

\begin{algorithm}[t!]
    \caption{Graph Constructor}
    \label{alg:graph_construction} 
    \begin{algorithmic}[1]
        \STATE \textbf{Input: } List of contexts \( C = [c_0, c_1, \ldots, c_n] \)
        \STATE Initialize empty graph \( G \gets \{\} \)
        \WHILE{C is not empty}
            \STATE \( c' \gets C[0] \); \( U \gets [] \)
            \FOR{each node \( c_i \) in \( C[1:] \)}
                \STATE \( r \gets \text{REL}(c_i, c') \)
                \STATE \( G \).add(\((c', c_i, r)\), \((c_i, c', r)\))
                \IF{\( r \) in ["counter", "dup"]}
                    \FOR{all \( (c', c_j, r_j) \)}
                        \STATE \( G \).add(\((c', c_i, r_j)\), \((c_i, c', r_j)\))
                    \ENDFOR
                \ELSE
                    \FOR{all \( (c', c_j, r_j) \)}
                        \IF{\( r_j \) in ["counter", "dup"]}
                            \STATE \( G \).add(\((c', c_i, r_j)\), \((c_i, c', r_j)\))
                        \ELSE
                            \STATE \( U \).append(\( c_i \))
                        \ENDIF
                    \ENDFOR
                \ENDIF
            \ENDFOR
            \STATE \( C \gets U \) (with duplicates removed)
        \ENDWHILE
        \STATE \textbf{Output: } fully connected graph \( G \)
\end{algorithmic}
\end{algorithm}

\paragraph{Reranker}

The \textit{Reranker} component organizes contexts into groups and removes unnecessary ones based on the constructed graph and solutions for each relationship type outlined in Section~\ref{sec4: solve}. First, for contexts in a distracting relationship, when a context with a descriptor is available, we remove the one without it. Next, for duplicated contexts, we select one randomly. Last, counterfactual contexts are separated into distinct groups, and the remaining contexts are distributed evenly across groups. When groups contain multiple contexts, the question is reformulated in a plural format. This systematic grouping aligns with relationship types, enhancing coherence and response accuracy.

\paragraph{Aggregator}
The \textit{Aggregator} component processes each group sequentially, generating responses and aggregating them with citations from the source contexts. This allows users to assess the origin of each response, offering transparency and supporting user judgment about the provided information. Since retrieval models may occasionally retrieve inaccurate information, and language models often struggle to verify document reliability, this approach gives users all relevant details with evidence, enabling them to make informed decisions.

\section{Experiments}
In this section, we share the experimental setup~(Section~\ref{sec5: setup}), six baselines~(Section~\ref{sec5: baseline}), experimental results~(Section~\ref{sec5: results}), and analysis over efficiency~(Section~\ref{sec5: flops}).

\subsection{Setup} \label{sec5: setup}
We evaluate various baselines across \ambigplus, \conflictqaplus, \ambig, and \conflictqa datasets, with eight models of varying sizes using the metric described in Section~\ref{sec3: metric}. We consider \textit{D-F1} as our primary metric as it captures the presence of the surface name, descriptor, and answer information, but we also report the performance of other metrics. Experiments are conducted on 1–4 A100 80GB GPUs at the models' maximum lengths. Following \citet{lee2024ambigdocs}, we select Llama2~\citep{touvron2023llama}, Mistral~\citep{jiang2023mistral}, ChatGPT, and additionally include recent models Llama3~\citep{dubey2024llama} and GPT-4o\footnote{reference to \href{https://openai.com/index/chatgpt/}{ChatGPT} and  \href{https://openai.com/index/hello-gpt-4o/}{GPT-4o}}. Refer to Appendix~\ref{app: model} for more details.

\subsection{Baselines}   \label{sec5: baseline}
We evaluate five baselines to assess how different context-handling strategies impact performance.  
\textit{Base} inputs all relevant contexts at once, resulting in long input lengths per question.  
\textit{Retrieve} and \textit{Summarize} also run in a single inference but aim to reduce input length. \textit{Retrieve} ranks contexts based on relevance and answer diversity, similar to \citet{min2021joint}, while \textit{Summarize} is inspired by \citet{xu2023recomp}, focusing on efficiency and performance by summarizing contexts. We use GPT-4 for both ranking and summarization.
\textit{Random} and \textit{KMeans} group contexts in the same count as \framework but differ in approach: \textit{Random} assigns groups randomly, while \textit{KMeans} clusters contexts via BERT embeddings~\citep{Devlin2019BERTPO}.  
\textit{Separate} processes each context individually. See Appendix~\ref{app: baselines} for more details of baselines.
In this study, we assume that the relevant paragraphs are pre-retrieved to remove the factor of retrieval error and focus specifically on generation ability. We leave the integration of retrieval from large corpora as future work. 

\begin{table*}[t!]
\centering
\fontsize{7.5}{10}\selectfont
    \begin{tabular}{ccc|ccc|ccc|c}
    \toprule
    \multicolumn{3}{c}{Number of Inference Run:}&\multicolumn{3}{c}{Single} & \multicolumn{3}{c}{Grouping} & \multicolumn{1}{c}{Individual}\\
    \midrule
   \multicolumn{3}{c}{Methods:}& Base& Retrieve & Summarize & Random & KMeans  & \framework & Separate \\
    \midrule
    \multirow{8}{*}{\ambigplus} 
    & \multirow{3}{*}{Llama2} 
    & 7B &17.0& 4.2 &18.8 & \underline{19.4} & 3.6 & \textbf{22.0} & 13.5\\
    && 13B & 16.4 &4.0 &\underline{17.2} &14.3 & 2.3  & \textbf{19.9}& 14.1 \\
    && 70B & 15.0 &11.2&15.5& 15.7 & 5.8  & \underline{17.9} & \textbf{18.7} \\
    \cmidrule{2-10}
    & \multirow{2}{*}{Llama3} 
    & 8B & 17.7 & 15.2&18.3&20.4 & 12.8 & \textbf{21.4}& \underline{20.5} \\
    && 70B & 10.6 &9.3&12.0& 11.7 & 8.5 & \underline{14.6}& \textbf{22.4} \\
    \cmidrule{2-10}
    & \multirow{1}{*}{Mistral} 
    & 7B & 22.9 & 18.2 &23.7& 24.0  & 14.1 & \underline{27.5}& \textbf{28.8}\\
    \cmidrule{2-10}
    & \multirow{2}{*}{GPT} 
    & ChatGPT & 19.2 & 17.9&24.3&21.6&10.6&\textbf{29.0}&\underline{27.3}\\
    && GPT-4o & 19.6 & 17.0&23.9&25.6&11.1&\underline{31.4}&\textbf{32.1}\\
    \midrule
    \multirow{8}{*}{\conflictqaplus} 
    & \multirow{3}{*}{Llama2} 
    & 7B & {27.8} & 5.8& 28.1 &\underline{29.1}  &19.3& \textbf{32.1} & 27.0\\
    && 13B & 30.3 & 1.5& 30.7&30.6 & 14.4 & \underline{30.8}& \textbf{37.5}\\
    && 70B & 26.7 & 9.1 & 29.2 & 17.0 & 9.5 & 31.4 & \textbf{40.1} \\
    \cmidrule{2-10}
    & \multirow{2}{*}{Llama3} 
    & 8B & 16.9 & 16.3& 17.7& 17.5  & 17.0& \underline{22.8}& \textbf{25.6} \\
    && 70B & 18.3 & 14.0& 17.9& 21.4 & 10.2 & \underline{28.7}& \textbf{35.2}\\
    \cmidrule{2-10}
    & \multirow{1}{*}{Mistral} 
    & 7B & 29.4 & 22.9&\underline{31.2}& 30.9  & 26.9 & \textbf{31.6}& 30.3\\
    \cmidrule{2-10}
    & \multirow{2}{*}{GPT} 
    & ChatGPT & 32.9 & 27.6& 31.2& 32.5&21.7 &\underline{35.9} & \textbf{37.1}\\
    && GPT-4o & 32.4 &22.3 &33.7& 35.1 & 23.0&\textbf{38.3} & \underline{38.1}\\    
    \bottomrule
    \end{tabular}
\caption
     {Overall Performance of \ambigplus and \conflictqaplus with DF-1. The best and second best of each model in \textbf{bold} and \underline{underline} respectively} 
\label{table: main_result}
\end{table*}

\subsection{Results}\label{sec5: results}

\begin{table}[t!]
\centering
\fontsize{7.5}{10}\selectfont
    \begin{tabular}{ccc|cccc}
    \toprule
     & & & Ans & Ent & EAR & D-F1 \\
    \midrule
    \multirow{7}{*}{7B}
    & \multirow{3}{*}{Single} 
    & Base & 51.4 & 39.5 & 25.7 & 17.0 \\
    && Retrieve & 44.1 & 16.3 & 7.9 & 4.2\\
    && Summarize & 60.0 & 42.1 &24.8 & 18.8\\
    \cmidrule{2-7}
    & \multirow{3}{*}{Grouping} 
    & Random &56.7 &35.8& 27.4& 19.4\\
    && KMeans & 42.6 & 14.2 & 6.3 & 3.6 \\
    && \framework & 61.4 & 41.2 & 34.0 & 22.0 \\
    \cmidrule{2-7}
    & \multirow{1}{*}{Individual}  
    & Separate &66.9 &41.6 &37.3 &13.5 \\
    \midrule
    \multirow{7}{*}{13B} 
    & \multirow{3}{*}{Single} 
    & Base& 48.8 & 31.3 & 19.3 & 16.4 \\
    && Retrieve &46.0& 14.9& 7.5&4.0 \\
    && Summarize &60.4 & 39.0 &21.9&17.2\\
    \cmidrule{2-7}
    & \multirow{3}{*}{Grouping} 
    & Random & 55.9& 25.6& 19.4& 14.3\\
    && KMeans & 40.1 &11.8 &4.0&2.3\\    
    && \framework & 58.3 & 39.3 & 29.4 & 19.9\\
    \cmidrule{2-7}
    & \multirow{1}{*}{Individual}  
    & Separate & 81.5& 48.1& 35.3& 14.1\\
    \midrule
    \multirow{7}{*}{70B} 
    & \multirow{3}{*}{Single} 
    & Base &60.5 & 37.1 & 27.5 & 15.0\\
    && Retrieve &50.0 & 27.9&20.2 &11.2 \\
    && Summarize &61.9 &31.6 &26.0& 15.5 \\
    \cmidrule{2-7}
    & \multirow{3}{*}{Grouping} 
    & Random &58.9 &26.1 &20.9 &15.7\\
    && KMeans & 38.6 &17.6 &8.0 &5.8\\
    && \framework & 63.8 & 37.2 & 31.7 & 17.9\\
    \cmidrule{2-7}
    & \multirow{1}{*}{Individual}  
    & Separate  & 89.0& 35.7& 27.3& 18.7\\
    \bottomrule
    \end{tabular}
\caption
     {Performance of Llama2 with various model size in \ambigplus with detailed metrics} 
\label{table: llama2_detail}
\end{table}

Table~\ref{table: main_result} shows the overall \textit{Disambig-F1 (D-F1)} across various models on the \ambigplus and \conflictqaplus datasets. 
Our results demonstrate that \framework consistently improves performance over six baselines across eight different models.
It achieves the highest performance among methods that use grouping-based inference and is comparable to Separate, which processes each context individually, though at a notably higher computational cost. Table~\ref{table: single_factor} in the Appendix shows a similar trend in \ambig and \conflictqa, datasets composed of contexts with single relationship.

Among the single-inference methods (Base, Retrieve, Summarize), Summarize achieves the highest performance due to its shorter input length that retains key details. However, it still struggles with handling complex relationships between contexts in a single inference run, resulting in lower performance than grouping-based methods.

Grouping-based methods (Random, KMeans, \framework) apply inference over context groups, underscoring the importance of the grouping strategy. Random grouping outperforms single-inference methods due to multiple inference runs, but lower performance than \framework. KMeans performs worse than Random, likely because clustering similar contexts makes it challenging for the model to generate answers with specific entity descriptors, reducing D-F1. Retrieve shows a similar trend to KMeans, as both methods group similar contexts together. Table~\ref{table: llama2_detail} shows that while both have high answer recall, they have particularly low entity recall. Such results emphasize the importance of the grouping method of \framework.

Separate, which processes each context individually, demonstrates performance comparable to or exceeding \framework but requires significantly more computation, as explored in the following section. Table~\ref{table: llama2_detail} highlights that Separate achieves high answer recall but lower entity recall, which we assume is due to processing single contexts without the broader exposure to varied descriptors. This lack of descriptor emphasis limits effective entity disambiguation, whereas methods processing multiple contexts expose models to various entities, supporting clearer differentiation.

Interestingly, larger models do not always outperform smaller ones. As shown in Table~\ref{table: llama2_detail}, while larger models generally achieve high answer recall, they often overlook descriptors, resulting in lower entity recall and overall disambiguation performance. However, generally more advanced language models, such as API and recent models, tend to exhibit stronger overall performance. See Appendix~\ref{app: factor} for more details.

\subsection{Efficiency} \label{sec5: flops}
\begin{figure}[t!]
    \centering
    \begin{minipage}[b]{0.45\textwidth}
    \includegraphics[width=\textwidth]{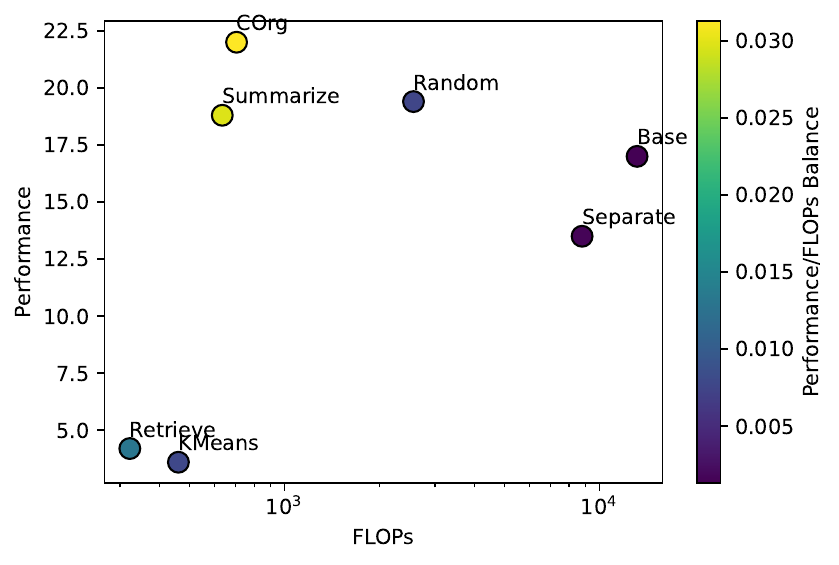}
    \end{minipage}
\caption{\fontsize{6.5}{10}\footnotesize Trade-off between efficiency and performance for Llama2-7B in \ambigplus. Light color indicates better balance between the two; \framework shows the best followed by Summarize. }
\vspace{-0.2em}
\label{fig: tradeoff}
\end{figure}

Figure~\ref{fig: tradeoff} shows the tradeoff between efficiency and performance for the Llama2-7B model on the \ambigplus dataset\footnote{Please note that FLOPs calculations here include only input-response processes, excluding additional steps like summarization, clustering, retrieval, or graph construction, as these are performed only once per dataset.}. \framework achieves the best balance, as it filters and groups contexts thus avoid processing all of them and have shorter responses. 
Summarize also performs efficiently by reducing input length through summarization. 
Retrieve and KMeans demonstrate the lowest FLOPs, as their outputs tend to be concise, containing only the answers. In contrast, Base and Separate are less efficient: Base uses long inputs and generates lengthy responses, while Separate produces responses for each context, increasing response length. For more details, see Appendix~\ref{app: eff}.

\section{Related Works}

Various studies highlight incorporating external knowledge as a solution for hallucination and expanding the model's capability. However, performance is dependent on the relationships among input contexts. Some works focus on ambiguous questions involving multiple relevant contexts~\citep{min-etal-2020-ambigqa, lee2024ambigdocs}, while others examine counterfactual or conflicting contexts relative to model knowledge~\citep{Chen2022RichKS, Longpre2021EntityBasedKC, lee2023well, Zhou2023ContextfaithfulPF, Xie2023AdaptiveCO}. Yet, real-world corpora often feature more complex relationships, which this paper categorizes into four types, revealing a gap in prior solutions for managing such complexities efficiently.

Some research leverages web-based corpora that might capture these relationships. However, we noticed that many of these works often rely on single sources, like Wikipedia, where each fact is likely to appear only once~\citep{lee2024ambigdocs, Longpre2021EntityBasedKC, min-etal-2020-ambigqa} or find relevant contexts using search engine API, which introduce bias to retrieve similar rather than diversely related contexts~\citep{yao2022react, schick2024toolformer, hao2024toolkengpt}. In this work, we focus on the real-world challenge of complex relationships among input contexts, and analyze the impact of each factor.

We further propose \framework, a framework that achieves high performance and efficiency across diverse real-world cases without additional training. Unlike previous works that use graph-based methods~\citep{edge2024local, Sarmah2024HybridRAGIK}, \framework constructs a graph where each node represents an entire document, and edges capture relationships between documents. In contrast, prior approaches focus on building graphs with nodes representing entities and edges capturing relationships within a single document. This distinction arises because our objective is to help LLMs better understand and process interconnected documents, whereas previous works focus on modeling relationships within individual documents. To the best of our knowledge, this is the first work to explore graph-based modeling for complex, interrelated documents.

\section{Conclusion}
This work investigates the characteristics of real-world corpora, categorizing them into four types and examining their effects and solutions. We find that no single solution addresses all four factors. Based on the observations, we introduce \ours~(\framework), a framework consisting of three components—graph constructor, reranker, and aggregator—designed to achieve high answer recall, effective disambiguation, and efficiency in real-world corpora with complex relationships between contexts. It is applicable to any model without requiring additional training. Experiments across four datasets with eight models demonstrate that \framework consistently enhances performance, outperforming six baselines and achieving the best trade-off between efficiency and effectiveness.

\section{Limitations}

Since our primary focus is on enhancing the language model's response accuracy, we control for retrieval model bias and error by pre-selecting relevant contexts for each question. This ensures that the answer always exists within the provided contexts. 

To maximize the knowledge available to users, we have the model generate all possible answers, including counterfactual ones, even if some may contain incorrect knowledge. Rather than having the model filter for correctness, we provide references to relevant documents, allowing users to make informed choices.

Additionally, given our goal to design a flexible pipeline that can easily adapt and benefit any newly released language model, we concentrate on optimizing performance at the inference stage instead of training new models. This approach supports broader applicability and immediate benefits with future models. We leave the exploration of training-based methods for handling complex interrelationships between answer contexts as future work.

\section*{Acknowledgements}
We thank Nishant Balepur, Vishakh Padmakumar, Dang Nguyen, Yoonjoo Lee, Zoey Ki, Paiheng Xu, and the people at Adobe Research for helpful discussions and constructive feedback.

\bibliography{anthology,custom}
\bibliographystyle{acl_natbib}

\appendix

\section{Complex, Interrelated Contexts} \label{app: sec2}

\subsection{Example of \ambigplus with each relationship}

\begin{table*}[t!]
\centering
\fontsize{7.5}{9}\selectfont
\caption{\fontsize{7.5}{10}\footnotesize Examples contexts containing each relationship given question ``2019 World Ice Hockey Championships host country?''} 
\begin{tabular}{ m{2cm} m{12cm} m{6cm}} 
    \toprule
    \textbf{Relationship} & \textbf{Context} \\
    \midrule
        & The 2019 IIHF Ice Hockey World Championship will be held in Slovakia, as confirmed by the IIHF on May 15, 2015. This marks the second occasion that Slovakia will host the championship as an independent nation. Similar to the 2011 event, the host cities will be Bratislava and Košice. The preliminary round seedings are determined by the 2018 IIHF World Ranking, following the conclusion of the 2018 IIHF World Championship. On May 22, 2018, the IIHF and the local organizing committee revealed the groups for the tournament.
         \\
        \midrule
        Duplicated 
        & The 2019 IIHF World Ice Hockey Championships were hosted by Slovakia. The tournament was the 83rd such event hosted by the International Ice Hockey Federation. Teams from around the world descended upon Slovakia to compete in the prestigious competition. The event took place in two cities, Bratislava and Kosice, from May 10 to May 26. This marked the second time that Slovakia has hosted the World Ice Hockey Championships, with the first occasion being in 2011. The tournament featured some of the best players from around the globe and was a significant event in the international hockey calendar
        \\
    \midrule
        Distracting
        & 2019 World Junior Ice Hockey Championships The 2019 IIHF World Junior Championship ("2019 WJHC") will be the 43rd Ice Hockey World Junior Championship. It will begin on December 26, 2018, and will end with the gold medal game being played on January 5, 2019. This will mark the 13th time that Canada has hosted the IHWJC. On December 1, 2016, it was announced that Vancouver and Victoria, British Columbia had won the bid to host the 2019 World Juniors. It will be the second time that Vancouver has been the primary host of the tournament and the first time that \\
        \midrule
        Counterfactual
         & The 2019 World Ice Hockey Championships organized by the International Ice Hockey Federation (IIHF) saw a joint hosting by Canada and British Columbia, specifically in the cities of Vancouver and Victoria. These cities were selected due to the excellent infrastructure and passionate fan base present there. The championships kicked off in the spring of 2019 and concluded later that year. It was a grand spectacle, with teams from all over the world competing for the prestigious title. This championship further solidified Vancouver and Victoria's reputations as prime locations for hosting international sporting events. \\
            \midrule
        Ambiguous
         & The 2019 World Ice Hockey Championships were hosted by Slovakia. Teams from around the world descended upon Slovakia to compete in the prestigious competition. The event took place in two cities, Bratislava and Kosice, from May 10 to May 26. This marked the second time that Slovakia has hosted the World Ice Hockey Championships, with the first occasion being in 2011. The tournament featured some of the best players from around the globe and was a significant event in the international hockey calendar. \\
    \bottomrule
\end{tabular}
\label{table: example_of_relationship}
\end{table*}

Examples in Table~\ref{table: example_of_relationship} illustrate the four types of relationships between contexts—ambiguous, counterfactual, duplicated, and distracting—when compared to a main context. In the ambiguous context, no descriptor is provided, whereas the main context includes the descriptor ``IIHF''. For the counterfactual context, while the main context states that the event was held in Slovakia, the counterfactual context claims it was jointly hosted in Canada and British Columbia. The duplicated context matches the main context across all information, including surface name, descriptor, and answer. In the distracting context, both the main and distracting contexts contain different descriptors—``IIHF'' for the main and ``junior'' for the distracting context.

\subsection{Human Evaluation of \ambigplus}

\begin{figure*}[t!]
{
\begin{tcolorbox}[width=0.99\textwidth, halign title=center, title = {Instruction to Human Evaluators}]

Definitions:
\newline
When given a context and a question, we can extract three things from the context. 
Answer: answer to the question from the given context
Entity: entity related to the answer in the simplest form
Descriptor: specific detail of the entity that distinguishes it from other entities. If not given, it is “Null”
For example, when given a context containing a sentence “A feature-length film, The Simpsons Movie, was released in theaters worldwide on July 27, 2007, to critical and commercial success, with a sequel in development as of 2018.” and question “When was The Simpsons released?”, the answer to the question is “July 27, 2007”, the entity is “The Simpsons” and the descriptor is “Movie”. When a sentence is “Since The Simpsons debut on December 17, 1989, 769 episodes of the show have been broadcast.”, the answer to the question is “December 17, 1989”, the entity is “The Simpsons”, and the descriptor is “Null” since there is no specific descriptor in the sentence.
\newline
We define whether the context is relevant to the question by:
* If the question has a descriptor, both the descriptor and entity of context should be same as question.
* If the question does not have a descriptor, only the entity of the context and the question should be the same.
\newline\newline
For those that the context is relevant to the question, we divide the document relationship into four and each case are defined as:\newline
* Ambiguous: different descriptor with either of the two being Null\newline
* Distracting: different descriptor with neither of the two being Null\newline
* Counterfactual: same descriptor with different answer\newline
* Duplicated: same descriptor with same answer\newline
For all cases, we consider that the entity is the same as in the question as the contexts are “relevant” to the question.
\newline\newline
** List of Contexts ** (We leave it empty since it is too long)
\newline\newline
Q1. Check if each context is relevant to the question. \newline
Question: 2019 World Ice Hockey Championships host country?\newline
Simply write “Yes” if you think it is relevant and “No” if you think it is not relevant.
\newline\newline
Q2. For the questions you consider relevant in Q1, provide an answer to the question for each context. For the irrelevant ones, please write “No”, and if the answer doesn’t exist in the context, please write “Null”. \newline
Question: 2019 World Ice Hockey Championships host country?
\newline	\newline
Q3. Identify which of the four defined document relationships (distracting, duplicated, counterfactual, ambiguous) exist within the corpus and write all that exists. Write the each relationship that exists in each cell.
\end{tcolorbox}
}
\caption{Instruction to Human Evaluators}
\label{fig: human_eval}
\end{figure*}

We recruited five freelancers using a platform to evaluate 10\% of randomly sampled examples from \ambigplus. The instructions, provided in Figure~\ref{fig: human_eval}, directed evaluators to answer three questions assessing dataset quality: (1) whether the generated contexts are relevant to the question, (2) if the answers are accurate and present within the document, and (3) whether the corpus captures the expected variety of real-world context relationships. To ensure that the freelancers understood the task, we divide the task into two step of first finishing 10\% of the task and checking before continuing over the rest. 

For Q1, 93.4\% of the contexts were considered relevant, with most irrelevance attributed to cases where the context contains too detailed descriptor compared to the question. 
For Q2, all answers are present within the provided contexts, and 89.0\% of human responses align with the original answer, as verified through GPT-4o to account for potential wording variations.
For Q3, 84.6\% of examples were rated as containing all four relationship factors, with 10.3\% having three factors and 5.1\% having fewer than three. 

Each multi-context question took an average of 7 minutes to evaluate. We compensated freelancers at an average rate of 350 dollars for completing 100 instances.

As \ambigplus and \conflictqaplus contain generated contexts, we acknowledge the potential risks involved. We asked the freelancers to check for any issues, but since all sources are drawn from publicly released datasets, they reported no findings.

\subsection{Statistics of Real-World Corpora} \label{app: stats}
To investigate the statistics of real-world corpora, we conduct an experiment using all questions in the AmbigDocs dataset. For each question, we retrieve the top 10 documents using the Bing API. On these retrieved documents, we apply the same procedure as the graph constructor (prompt in Figure~\ref{table: app_ambiguous}) to calculate the statistics for each relationship type. Specifically, for each question, we determine the composition rate of each relationship type and then average these rates across all questions. 

To evaluate the diversity of answers captured by the AmbigDocs questions (32.7\%), we compute the answer recall across all retrieved documents with a list of annotated answers in AmbigDocs. For example, if a question has five annotated answers, we check whether each answer is present in any of the retrieved documents. If at least one document contains a given answer, we consider that answer as covered. The coverage rate for each question is calculated as the number of answers covered divided by the total number of annotated answers. 

\subsection{Details of corpus construction}

\begin{figure*}[t!]
{
\begin{tcolorbox}[width=0.99\textwidth, halign title=center, title = {Prompt to Generate Context}]

Given a question and an answer, generate an evidence context consisting of 6-7 sentences. The purpose of the context is for people to read and answer the question. The answer and information in the context do not have to be true.\newline\newline  
Question: how many episodes are in chicago fire in season 4?\newline Answer: 103\newline Context: The fourth season of Chicago Fire , an American drama television series with executive producer Dick Wolf , and producers Derek Haas , Michael Brandt , and Matt Olmstead , was ordered on February 5 , 2015 , by NBC , and premiered on October 13 , 2015 and concluded on May 17 , 2016 . The season contained 103 episodes .\newline\newline Question: how many episodes are in chicago fire?\newline Answer: 103\newline Context: Chicago Fire is a gripping American drama television series comprised of 103 episodes that delves into the lives of the firefighters, rescue personnel, and paramedics of Firehouse 51 of the Chicago Fire Department. The series offers an inside look at the professional and personal challenges faced by these brave men and women as they risk their lives to save others. The show captures the intense camaraderie, complex relationships, and high-stakes situations that define their everyday existence. With a compelling mix of action, drama, and emotional depth, Chicago Fire provides an authentic and engaging portrayal of life on the front lines of emergency response.
\end{tcolorbox}
}
\caption{Prompt to Generate Context}
\label{fig: generate_context}
\end{figure*}

\begin{figure*}[t!]
{
\begin{tcolorbox}[width=0.99\textwidth, halign title=center, title = {Prompt to Generate Sub-Questions for ConflictQA}]
Given a question and list of answers, generate a detailed question for each of the given answer should be answer to the question respectively. Please note that the number of given answers should be the same with generated detailed questions and the answer should NOT be in the generated question.\newline
---\newline
question: who proposed evolution as the basis of biological development?\newline answers: Jodie Foster, Mara Jade, Yeh Raaste Hain Pyaar Ke, Billy Joel, Rigg\newline\newline detailed questions: who proposed evolution in 1859 as the basis of biological development? // who proposed evolution in 1863 as the basis of biological development? // who proposed evolution as the basis of biological development in 1871? // who proposed evolution as the basis of biological development in 1921? // who proposed evolution as the basis of biological development in 1951?\newline ---\newline question: who sings gim me some lovin in days of thunder?\newline answers: AB de Villiers, UMBC, Nashville Predators\newline\newline detailed questions: who first sings gim me some lovin in days of thunder? // who remake gim me some lovin in days of thunder? // who sings gim me some lovin in days of thunder part 2?\newline ---\newline question: how many episodes of grey anatomy?\newline answers: 501, 216\newline\newline detailed questions: how many episodes of greys anatomy season 14? // how many episodes of greys anatomy season 12?\newline ---
\end{tcolorbox}
}
\caption{Prompt to Generate Sub-Questions for ConflictQA}
\label{fig: generate_sub_query}
\end{figure*}

To evaluate the LLM's ability with corpora representing real-world scenarios, we generate additional contexts based on existing datasets consisting of (question, answer, contexts) pairs. 
We use \ambig~\citep{lee2024ambigdocs}, which contains distracting contexts, and dataset released from \citet{Zhou2023ContextfaithfulPF}, which includes counterfactual contexts as shown in Table~\ref{table: dataset}. For simplicity, we name the dataset released from \citet{Zhou2023ContextfaithfulPF} as \conflictqa.
Based on the information provided in each dataset, we add ambiguous, conflicting, and duplicated contexts to \ambig, and add distracting, ambiguous, and duplicated contexts to \conflictqa, which we name as \ambigplus and \conflictqaplus, respectively. 

In detail, \ambig contains pairs of question $q$ and a list of contexts relevant to the question $C_q$. The question $q$ asks about an entity with a surface name but without a descriptor. The list $C_q$ contains $N$ pairs of sub-questions $q'_i$ and answer $a_i$ where the sub-question asks about the entity with the same surface name but with descriptor and the answer $a_i$ is the answer to both question $q$ and $q'_i$ ($\{(q'_i, a_i) \mid i = 0, \ldots, N\}$).
For each question $q$, to add counterfactual contexts in $C_q$, we randomly select a sub-question $q'_i$ and for each possible answer from the list except for the correct one, we instruct GPT-4 to generate contexts that could produce the correct answer given the sub-question and answer ($\text{Counterfactual contexts} = \{\text{LLM}(q'_i, a_j) \mid j = 0, \ldots, N, \, j \neq i\}$). 
For the duplicated case, we provided the model with sub-question and answer pairs to create matching contexts ($\text{Duplicated contexts} = \{\text{LLM}(q'_j, a_j) \mid j = 0, \ldots, N\}$). 
For the ambiguous case, we provide the model with question and answer ($\text{Ambiguous contexts} = \{\text{LLM}(q, a_j) \mid j = 0, \ldots, N\}$).

ConflictQA contains pairs of question $q$\footnote{Most questions do not include a descriptor, but those that do were revised by humans to remove it.} along with a list of relevant contexts $C_q$. Unlike \ambig, the contexts in $C_q$ are counterfactual to each other and often lack descriptors. 
To align with \ambig's structure, we use GPT-4 to generate plausible sub-questions with descriptors for each answer in $C_q$ (prompt in Figure~\ref{fig: generate_sub_query}). 
Using these generated sub-questions, we add distracting contexts to $C_q$ by generating a context for each sub-question and answer ($\text{Distracting contexts} = \{\text{LLM}(q'_j, a_j) \mid j = 0, \ldots, N\}$).
For the duplicated and ambiguous contexts, we follow the same process as in \ambig. Figure~\ref{fig: generate_context} shows the prompt used to generate contexts.

\subsection{Statistics of Dataset}
\begin{figure}[t!]
    \centering
    \begin{minipage}[b]{0.45\textwidth}
    \includegraphics[width=\textwidth]{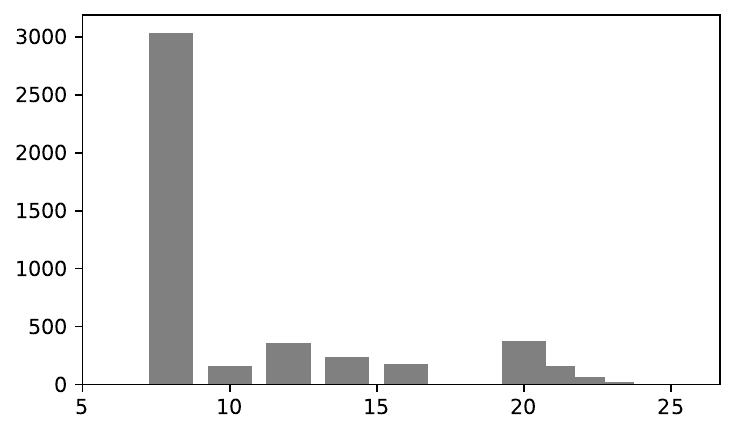}
    \end{minipage}
\caption{\fontsize{6.5}{10}\footnotesize Statistics showing the number of questions (y-axis) against the number of contexts per question (x-axis). Most cases have eight contexts for each question.}
\label{fig: doc_cnt}
\end{figure}

Figure~\ref{fig: doc_cnt} illustrates the number of contexts for each question. Typically, there are eight contexts per question, with a maximum of 25 contexts observed.
The \ambig and \ambigplus datasets each contain 6,000 evaluation samples, while the \conflictqa and \conflictqaplus datasets include 1,000 evaluation samples.

\section{Analyzing Solution for each Factor}

\subsection{Ablation Studies on Distracting and Ambiguous Relationships}

Combining pluralizing the question with an additional prompt (Plural \& Extra Prompt) generally leads to further improvements in model performance. Additionally, separating all contexts (Separation) indicates a significant enhancement, particularly in answer recall. However, processing all contexts separately tends to be computationally intensive making it impractical.

\begin{table}[t!]
\centering
\fontsize{8.5}{10}\selectfont
    \begin{tabular}{c|cccc}
    \toprule
    & Ent & Ans & EAR & D-F1 \\
    \midrule
    & 52.5 & 53.0 & 36.8 & 21.7 \\
    One Shot & 48.4 & 61.4 & 37.6 & 18.2 \\
    Extra Prompt & 53.0 &56.3& 38.3 & 27.6 \\
    \textbf{Plural} & 68.9 & 67.9 & 42.7 & 28.0 \\
    \midrule
    Plural \& Extra Prompt & 65.6 &72.0& 51.5 & 30.6 \\
    Separation & 64.7 & 84.7 & 54.5 & 30.4 \\
    \bottomrule
    \end{tabular}
\caption
     {\fontsize{8.5}{10}\footnotesize Performance of Llama2-7B using only contexts with distracting relationships.} 
\label{table: app_distracting}
\end{table}

\begin{table}[t!]
\centering
\fontsize{8.5}{10}\selectfont
    \begin{tabular}{c|cccc}
    \toprule
    & Ent & Ans & EAR & D-F1 \\
    \midrule
    & 31.0 & 49.6 & 29.9 & 17.2 \\
    One Shot & 34.4 & 52.2 & 32.0 & 18.3 \\
    Extra Prompt & 40.7 & 50.9 & 31.5 & 18.4 \\
    Plural & 45.1 & 51.9 & 33.2 & 20.1 \\
    \textbf{Change to Distracting} & \text{52.5} & \text{53.0} & \text{36.8} & \text{21.7} \\
    \midrule
    w/ plural and Extra Prompt &51.6 &77.0& 40.5 & 26.6 \\
    Separation & 55.7 & 85.6 & 45.0 & 28.4 \\
   
    \bottomrule
    \end{tabular}
\caption
     {\fontsize{8.5}{10}\footnotesize Performance of Llama2-7B using only contexts with ambiguous relationships.} 
\label{table: app_ambiguous}
\end{table}

\section{\ours} \label{app: ours}
\begin{figure*}[t!]
{
\begin{tcolorbox}[width=0.99\textwidth, halign title=center, title = {Input Format for Graph Constructor}]

When given a context and a question, we can extract three things from the context. Answer: answer to the question from the given context Entity: entity related to the answer in the simplest form Descriptor: specific detail of the entity that distinguishes it from other entities. If not given, it is “Null” For example, when given a context containing a sentence “A feature-length film, The Simpsons Movie, was released in theaters worldwide on July 27, 2007, to critical and commercial success, with a sequel in development as of 2018.” and question “When was The Simpsons released?”, the answer to the question is “July 27, 2007”, the entity is “The Simpsons” and the descriptor is “Movie”. When a sentence is “Since The Simpsons debut on December 17, 1989, 769 episodes of the show have been broadcast.”, the answer to the question is “December 17, 1989”, the entity is “The Simpsons”, and the descriptor is “Null” since there is no specific descriptor in the sentence. \newline\newline For those that the context is relevant to the question, we divide the document relationship into four and each case are defined as: \newline* Ambiguous: different descriptor with either of the two being Null and with same answer \newline* None: different descriptor with either of the two being Null and with different answer \newline* Distracting: different descriptor with neither of the two being Null \newline* Counterfactual: same descriptor with different answer \newline* Duplicated: same descriptor with same answer\newline When given contexts, could you generate relation from Context1 to rest of the contexts? \newline\newline---- \newline\newline [Context1] Title: 2019 IIHF World Championship Text: The 2019 World Ice Hockey Championships were notably held in Canada, specifically in the province of British Columbia. The event took place in the vibrant cities of Vancouver and Victoria, ... The choice of Canada, a nation with a rich hockey heritage, underscored the significance of the tournament. By hosting the championships in Vancouver and Victoria, Canada once again demonstrated its central role in the world of ice hockey. \newline [Context2] Title: 2019 IIHF World Championship Text: 2019 IIHF World Championship The 2019 IIHF Ice Hockey World Championship is scheduled to be hosted by Slovakia, as announced by the IIHF on 15 May ... The seedings in the preliminary round are based on the 2018 IIHF World Ranking, as of the end of the 2018 IIHF World Championship. On 22 May 2018, the IIHF and the local organizing committee announced the groups, in which \newline[Context3] Title: 2019 IIHF World Championship Text: The 2019 World Ice Hockey Championships were held in two different countries, France and Hungary. These nations played a significant role in hosting the event, ... and manage a large-scale sporting event. This joint hosting effort helped promote the sport within their borders and offered a memorable experience for all participants. \newline Question: 2019 World Ice Hockey Championships host country? \newline Relations: \newline Context2 - None \newline Context3 - Counterfactual \newline\newline
\end{tcolorbox}
}
\caption{Input format to graph constructor}
\label{fig: graph_constructor}
\end{figure*}

\subsection{Graph Constructor}
Figure~\ref{fig: graph_constructor} shows the prompt to GPT-4 to extract the relationship between contexts thus constructing a graph.

When constructing a graph, we classify \textit{ambiguous} relationships into two types: contexts that contain the same information as another context thus making them interchangeable, and those that do not. This classification allows us to apply the solution outlined in Section~\ref{sec4: solve}. Contexts that are interchangeable are labeled as having an ``ambiguous'' relationship, while those with no shared information are classified as having a ``None'' relationship, indicating no connection between them. Since a single context can relate to multiple contexts, the ``None'' relationships can often be disregarded and processed alongside other relationships beforehand.

\section{Experiments} \label{app: exp}
\subsection{Model Choice} \label{app: model}
We utilized models released on HuggingFace for our experiments. Below, we provide the detailed links and versions of the models used. 

\begin{itemize}
    \item Llama2: \href{https://huggingface.co/meta-llama/Llama-2-7b-chat-hf}{meta-llama/Llama-2-7b-chat-hf}, \href{https://huggingface.co/meta-llama/Llama-2-13b-chat-hf}{meta-llama/Llama-2-13b-chat-hf}, \href{https://huggingface.co/meta-llama/Llama-2-70b-chat-hf}{meta-llama/Llama-2-70b-chat-hf}
    \item Llama3: \href{https://huggingface.co/meta-llama/Llama-3.1-8B-Instruct}{meta-llama/Llama-3.1-8B-Instruct}, \href{https://huggingface.co/meta-llama/Llama-3.1-70B-Instruct}{meta-llama/Llama-3.1-70B-Instruct}
    \item Mistral: \href{https://huggingface.co/mistralai/Mistral-7B-Instruct-v0.2}{mistralai/Mistral-7B-Instruct-v0.2}
    \item ChatGPT: \href{https://platform.openai.com/docs/models/gpt-3.5-turbo}{gpt-3.5-turbo}
    \item GPT-4o: \href{https://platform.openai.com/docs/models/gpt-4o}{gpt-4o}
    \item GPT-4: \href{https://platform.openai.com/docs/models/gpt-4}{gpt-4}
\end{itemize}

\subsection{Baselines}  \label{app: baselines}
We evaluate over five baselines to evaluate how different approaches to operating contexts impact performance.
\textit{Base} represents a model that inputs all relevant contexts simultaneously, resulting in long input lengths for each question. 
The \textit{Retrieve} method employs GPT-4 to rank contexts based on not only similarity but also diversity, inspired by the work of \citet{min2021joint}. This approach aims to reduce the number of contexts by removing unnecessary ones and executing a single inference run, following \citet{lee2024can}. We utilize the top five ranked contexts for retrieval\footnote{As the model and code for \citet{min2021joint} are not available, and studies have shown that long-context models perform well in retrieval tasks~\citep{lee2024can}, we use GPT-4 for ranking.}.
The \textit{Summarize} approach builds on the idea of summarizing input contexts to enhance efficiency and maintain sufficient performance~\citep{xu2023recomp}. We utilize GPT-4 to summarize the relevant contexts, which are then employed to generate responses.
Both \textit{Random} and \textit{KMeans} group contexts into the same number of groups as \framework, though they differ from \framework in grouping methods: \textit{Random} groups contexts randomly, while \textit{KMeans} clusters contexts based on BERT embeddings~\citep{Devlin2019BERTPO}. This setup ensures the same group counts, isolating the effect of the grouping technique itself. 
\textit{Separate} treats each context individually, processing one per inference run and then concatenating all outputs. While this method reduces input length per run, it increases overall output length and time cost as it requires multiple runs.
In this study, we assume that the relevant paragraphs for each question are pre-retrieved to remove the factor of retrieval error and focus specifically on how the language model’s generation can be improved. We leave the integration of retrieval from large corpora with generation modeling as future work.

\begin{table*}[t!]
\centering
\fontsize{7.5}{10}\selectfont
    \begin{tabular}{ccc|ccc|ccc|c}
    \toprule
    \multicolumn{3}{c}{Number of Inference Run:}&\multicolumn{3}{c}{Single} & \multicolumn{3}{c}{Grouping} & \multicolumn{1}{c}{Individual}\\
    \midrule
   \multicolumn{3}{c}{Methods:}& Base& Retrieve & Summarize & Random & KMeans  & \framework & Separate \\
    \midrule
    \multirow{8}{*}{\ambig} 
    & \multirow{3}{*}{Llama2} 
    & 7B &20.0 & 4.8 &\underline{20.1} &19.4 &5.2 &\textbf{22.0} & 13.5 \\
    && 13B & 18.4 & 5.1&\underline{19.2}&14.3 &3.8 &\textbf{19.9}&12.1 \\
    && 70B & 17.0 &12.9 &15.9& 5.7 &10.2 &17.9 & 18.7 \\
    \cmidrule{2-10}
    & \multirow{2}{*}{Llama3} 
    & 8B & 19.7 & 14.5&20.0&\textbf{20.9}&11.4&\textbf{21.4}&20.5\\
    && 70B & 10.6&13.7&14.2&11.7&12.1&\underline{14.6}&\textbf{22.4} \\
    \cmidrule{2-10}
    & \multirow{1}{*}{Mistral} 
    & 7B & 25.9 & 23.7&24.8&24.0&14.9&\underline{27.5}&\textbf{28.8}\\
    \cmidrule{2-10}
    & \multirow{2}{*}{GPT} 
    & ChatGPT &19.2 & 15.6 &19.8 & \underline{21.2} & 14.8 & 20.4 & \textbf{23.1} \\
    && GPT-4o & 20.1 & 19.2 & 20.9 & 21.8 & 18.4 & \underline{23.0} & \textbf{25.9}\\
    \midrule
    \multirow{8}{*}{\conflictqa} 
    & \multirow{3}{*}{Llama2} 
    & 7B & 12.7 & 13.1 & 9.2 & 13.2 & 7.9 & \underline{14.3} & \textbf{16.0}\\
    && 13B & 10.7 & 10.4 & 9.8 & \underline{10.9} & 8.4 & 10.6 & \textbf{13.7}\\
    && 70B &  10.2 & 11.7 & 8.1 & 11.4 & 8.0 & \underline{12.0} & \textbf{15.3}\\
    \cmidrule{2-10}
    & \multirow{2}{*}{Llama3} 
    & 8B & 11.4 & 10.7 & 10.0 & 12.8 & 12.1 & \textbf{17.3} & \underline{16.8}\\
    && 70B & 9.8 & 8.8 & 5.7 & 12.5 & 10.3 & \underline{16.3} & \textbf{18.9}\\
    \cmidrule{2-10}
    & \multirow{1}{*}{Mistral} 
    & 7B & 12.3 & 10.2 & 9.4 & 14.0 & 12.7 & \underline{16.9} & \textbf{19.4}\\
    \cmidrule{2-10}
    & \multirow{2}{*}{GPT} 
    & ChatGPT & 14.7 & 13.9 & 10.8 & 15.3 & 11.6 & \underline{16.2} & \textbf{17.9}\\
    && GPT-4o & 16.3 & 12.7 & 10.5 & \underline{16.9} & 13.2 & 15.8 & \textbf{20.1}\\    
    \bottomrule
    \end{tabular}
\caption
     {Overall Performance of \ambig and \conflictqa with DF-1. The best and second best of each model in \textbf{bold} and \underline{underline} respectively.} 
\label{table: single_factor}
\end{table*}

\subsection{Results} \label{app: factor}

\paragraph{Performance on Corpora with Single-Factor Relationships}
Table~\ref{table: single_factor} presents the performance of \framework and baseline models on the \ambig and \conflictqa datasets, which consist of corpora containing only distracting and counterfactual relationships, respectively. We observe that for both datasets, \framework generally enhances performance, similar to the findings in Table~\ref{table: main_result}. However, the benefits of using \framework are more pronounced when the corpus exhibits more complex relationships. Additionally, in the case of \conflictqa, Separate demonstrates consistently strong performance, in line with the observations discussed in Section~\ref{sec4: solve}.

\paragraph{Effect of each factor}
\begin{table}[t!]
\centering
\fontsize{8.5}{10}\selectfont
    \begin{tabular}{l|c}
    \toprule
\textbf{Variants}                                                  & \textbf{Score} \\ 
\midrule
COrg                                                           & 22.0           \\ 
(1) without removing duplicates [R]                 & 21.6           \\
(2) without converting distracting to ambiguous [R]   & 20.2           \\
(3) without grouping    [R]                          & 19.1           \\ 
(4) without query reformulation in aggregator  [A]            & 19.9           \\ 
(5) without all                                            [R,A] & 17.0           \\ 
\bottomrule
\end{tabular}
\caption{Impact of removing different components from the reranker and aggregator. [R] indicates changes in the reranker and [A] indicates changes in the aggregator.}
\label{table:each_factor}
\end{table}

 Table~\ref{table:each_factor} shows the impact of removing different components from the reranker and aggregator\footnote{Please note that without the graph constructor, as we cannot process both the reranker and aggregator, we are not able to perform ablation over the graph constructor.}. For the reranker, we evaluate the effect of removing individual factors processed in the reranker (1), (2), and (3) to understand their individual contributions. For the aggregator, we investigate the removal of query reformulation (4) to evaluate its importance. Due to time constraints, we conduct experiment with AmbigDocs+ dataset with the Llama2-7b-chat model. Our experiment in the below table shows that removing the grouping (3) resulted in a significant performance drop; when documents with counterfactual relationships were processed together without any grouping. Similarly, removing query reformulation (4) tends to cause a noticeable decrease in performance. On the other hand, removing duplicate documents led to only a minor performance change, though it would increase input token consumption, as all context had to be processed. Overall, these findings highlight the importance of each component in the \framework framework. We will include these results in the final version of the paper. Additionally, we would like to emphasize the novelty of our grouping method using the graph constructor, as demonstrated by the comparison with other grouping methods (Random and KMeans).

\begin{table*}[t!]
\centering
\fontsize{8.5}{10}\selectfont
    \begin{tabular}{c|ccc|ccc|c}
    \toprule
    & Base (S) & Retrieve (S) & Summarize (S) & Random (G) & KMeans (G) & COrg (G) & Separate (I) \\
    \midrule
    \textbf{Input Token Count}  & All Context       & 773.85               & 354.63               & All Context         & All Context        & 752.18           & All Context       \\ 
    \textbf{Output Token Count} & 87.91             & 22.97               & 28.18               & 121.77             & 66.29             & 35.02            & 210.78           \\ 
    \textbf{Performance}        & 17.0              & 4.2                 & 18.8                & 19.4               & 3.6               & 22.0             & 13.5             \\ \bottomrule
\end{tabular}
\caption{Average token consumption for both input and output when generating with Llama-2-7b on \ambigplus. (S) means "S"ingle inference run, (G) means "G"rouping inference run, and (I) means "I"ndividual runs.}
\label{table:token_count}
\end{table*}

\subsection{Efficiency} \label{app: eff}

\paragraph{Statics of groups}
We investigate the number of groups, or inference runs, for Llama2-7B in the \ambigplus dataset. We observe that the majority of cases are grouped into two (3.5k), with additional instances of a single run (1.2k) and three runs (1.0k), along with a few cases featuring four or more runs (0.2k). This result aligns with Figure~\ref{fig: doc_cnt}, which shows that most cases consist of eight contexts. If each relationship is evenly divided, this configuration would yield two counterfactual contexts.

\paragraph{Average Token Consumption}
Table~\ref{table:token_count} shows the average token consumption for both input and output when generating with Llama-2-7b on AmbigDocs+. 
We could see the \framework tends to show a short input sequence as we do not process the entire context, which requires 1347.9 tokens for input. Instead, we remove unnecessary or redundant context, such as duplicated information or relationships that are not relevant (e.g., changing distracting relationships to ambiguous ones) during the reranker process. Ours also show the shortest output length compared to other group-based methods, those with (G), while maintaining high performance. Upon reviewing the outputs, we observed that COrg generates shorter responses as they focus only on the relevant information without appending unnecessary knowledge to the question or their parametric knowledge. Also, in the case of retrieve, we could see that it is especially short as they tend to generate responses without containing multiple answers as they can not answer the question. While with low token consumption, \framework shows the best performance, which further supports the practical efficiency of \framework in real-world deployment.

\end{document}